\documentclass[manuscript]{acmart}

\AtBeginDocument{%
  \providecommand\BibTeX{{%
    \normalfont B\kern-0.5em{\scshape i\kern-0.25em b}\kern-0.8em\TeX}}}

\setcopyright{acmcopyright}
\copyrightyear{2018}
\acmYear{2023}
\acmDOI{XXXXXXX.XXXXXXX}

\acmConference[Conference acronym 'XX]{Make sure to enter the correct
  conference title from your rights confirmation emai}{June 03--05,
  2018}{Woodstock, NY}
\acmPrice{15.00}
\acmISBN{978-1-4503-XXXX-X/18/06}

\usepackage{comment}
\usepackage{amssymb} 

\usepackage{color}

\newcommand{\figref}[1]{Fig.~\ref{fig:#1}}
\newcommand{\tabref}[1]{Table~\ref{tbl:#1}}

\newcommand{\overviewplus}{
 \begin{figure}[t]
    \centering
     \includegraphics[width=1.0\linewidth]{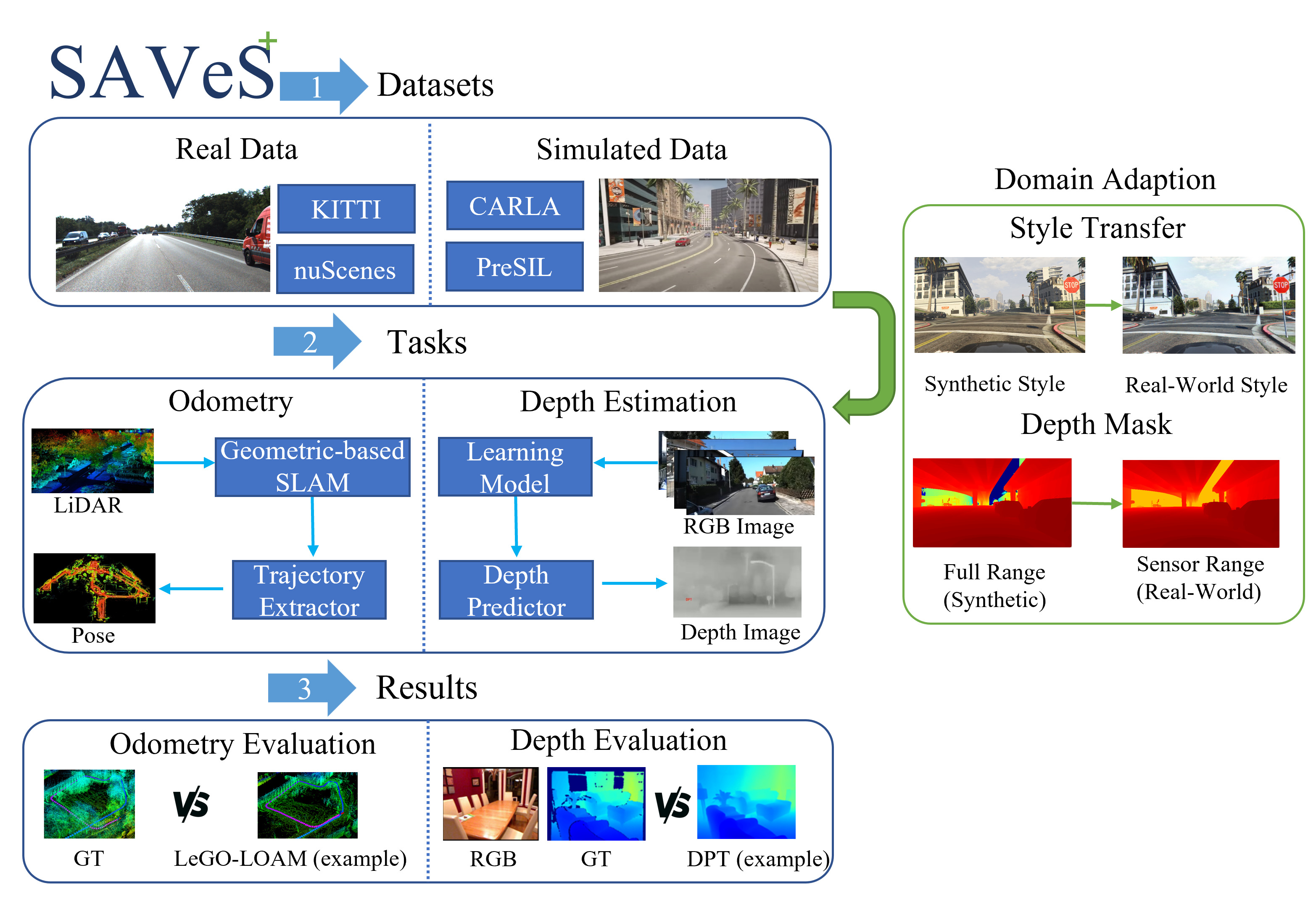}
  \caption{An overview of the Scoping Autonomous Vehicle Simulation (SAVeS) platform with domain adaption capabilities (SAVeS$^{+}$) for evaluating and improving mobility performance in simulated environments targeted for autonomous ground vehicle testing. The green arrow indicates that the plugin works on simulated data for depth task.} 
     \label{fig:overviewplus}
 \end{figure}
}

\newcommand{\quality}{
 \begin{figure}[t]
    \centering
     \includegraphics[width=1\linewidth]{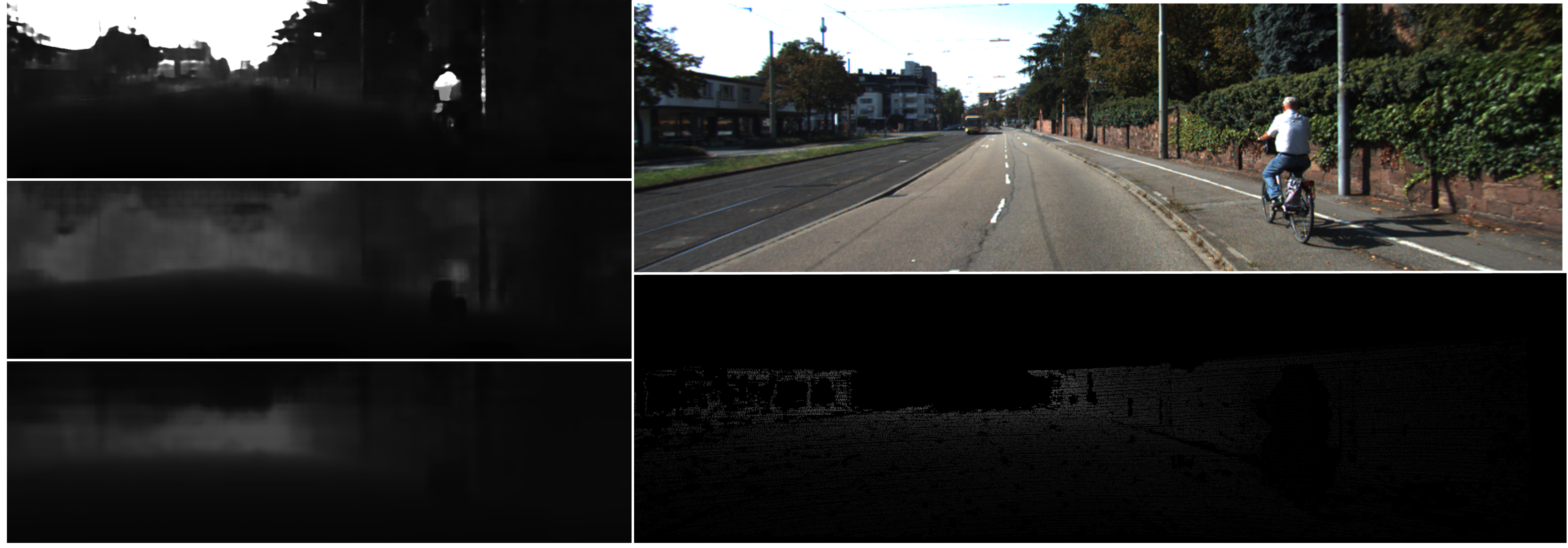}
  \caption{Qualitative estimated depth map from BTS trained on aforementioned three datasets: PreSIL (upper left), SAVeS+PreSIL (middle left) and KITTI (lower left), as well as RGB image (upper right) and sparse ground truth (lower right).} 
     \label{fig:quality}
 \end{figure}
}

\usepackage{multirow}
\usepackage{array}

\newcommand{\depth}{
\begin{table*}[t]
\caption{Learning-based algorithms tested on real and simulated environments. $y_i$ represents the $i$th valid pixel's depth ground truth in an image, while $y_i^*$ denotes the corresponding predicted depth. $n$ is the number of valid pixels, with the conditions $y_i > 0$ and $y_i^* > 0$. The difference, $d_i$, is calculated as $d_i = \log y_i - \log y_i^*$. For each image, evaluation metrics are computed as $D(y, y^*)$ based on the listed equations, and the mean of all images involved is calculated. For more detailed information, please see \cite{kitti}. The SILog metric is used as the primary factor to represent depth distribution without range disturbances.} 
\footnotesize
\begin{center}
\begin{tabular}{l l l |c c c c}
 &  &
\multicolumn{4}{c}{\textbf{Evaluate Metrics}}\\
\cline{4-7}
& & &SILog $\downarrow$ &sqErrorRel (\%)   $\downarrow$ &absErrorRel (\%) $\downarrow$ &iRMSE (1$/$km) $\downarrow$\\ 
\textbf{Dataset} & \textbf{Domain} & \textbf{Model} & $\dfrac{1}{n}\sum_{i}d_i^2-\dfrac{1}{n^2}(\sum_{i}dA_i)^2$ & $\sqrt{\dfrac{1}{n}\sum_{i}(\dfrac{y_i-y_i^*}{y_i})^2}$ & $\dfrac{1}{n}\sum_{i}\dfrac{|y_i-y_i^*|}{y_i}$ & $\sqrt{\dfrac{1}{n}\sum_{i}(\dfrac{1}{y_i}-\dfrac{1}{y_i^*})^2}$
\\ \hline

\multirow{3}{*}{KITTI}& \multirow{3}{*}{Real-World}& \multicolumn{1}{|l|}{GeoNet} & \textbf{0.1503$\pm$0.0268} & \textbf{0.0279$\pm$0.0097} & \textbf{0.1074$\pm$0.0210} & \textbf{0.0130$\pm$0.0032}\\ 
&&\multicolumn{1}{|l|}{DPT} &0.1926$\pm$0.0107 &0.0793$\pm$0.1439 &0.1557$\pm$0.0134 &0.0165$\pm$0.0005 \\
&&\multicolumn{1}{|l|}{AdaBins} &0.3861$\pm$0.0257  &0.1502$\pm$0.0243 &0.2923$\pm$0.0217 &0.0330$\pm$0.0005\\  \hline

\multirow{3}{*}{nuScenes}&\multirow{3}{*}{Real-World}& \multicolumn{1}{|l|}{GeoNet} &0.3527$\pm$0.0079  &0.1278$\pm$0.0085 &0.2527$\pm$0.0051 &0.0278$\pm$0.0002\\ 
&&\multicolumn{1}{|l|}{DPT} &\textbf{0.2635$\pm$0.0037}  &\textbf{0.1154$\pm$0.0136} &\textbf{0.1805$\pm$0.0018} &\textbf{0.0191$\pm$0.0010}\\ 
&&\multicolumn{1}{|l|}{AdaBins} &0.5667$\pm$0.0085  &0.3579$\pm$0.0493 &0.4574$\pm$0.0250 &0.0493$\pm$0.0009\\  \hline 

\multirow{3}{*}{CARLA}&\multirow{3}{*}{Synthetic}& \multicolumn{1}{|l|}{GeoNet} & 0.3819$\pm$0.0180 &  0.1933$\pm$0.0622& 0.3314$\pm$0.0527& 0.0182$\pm$0.0013\\ 
&&\multicolumn{1}{|l|}{DPT} &\textbf{0.2945$\pm$0.0202}  &\textbf{0.1520$\pm$0.0335} &\textbf{0.2418$\pm$0.0302} &\textbf{0.0140$\pm$0.0017}\\ 
&&\multicolumn{1}{|l|}{AdaBins} &0.5162$\pm$0.0361  &0.4666$\pm$0.0775 &0.5059$\pm$0.0322 &0.0262$\pm$0.0013\\  \hline 

\multirow{3}{*}{PreSIL}& \multirow{3}{*}{Synthetic}&\multicolumn{1}{|l|}{GeoNet} & 0.4086$\pm$0.0116 &0.1684$\pm$0.0117 &0.2479$\pm$0.0080 &0.0697$\pm$0.0006\\ 
&&\multicolumn{1}{|l|}{DPT} &\textbf{0.3152$\pm$0.0130}  &\textbf{0.0849$\pm$0.0050} &\textbf{0.1696$\pm$0.0076} &\textbf{0.0510$\pm$0.0329}\\ 
&&\multicolumn{1}{|l|}{AdaBINS} &0.6575$\pm$0.0169  &0.3838$\pm$0.0219 &0.4558$\pm$0.0140 &0.0694$\pm$0.0013\\  \hline 

\end{tabular}
\label{tbl:depth}
\end{center}
\end{table*}
}
\newcommand{\geometric}{
\begin{table}[t]
\caption{Geometric-based algorithms tested on real and stimulated environments. $\|trans(\cdot)\|$ stands for Euclidean Distance. For APE, at any timestamp i, $E_i=P_{est,i} \ominus P_{ref,i}=P_{ref,i}^{-1}P_{est,i}$
where $P_{ref,i}$ and $P_{est,i}$ are poses, and $\ominus$ is invert positional operator to calculate relative pose. For RPE, $E_{i,j}=\delta_{est_{i,j}}\ominus\delta_{ref_{i,j}}=(P_{ref,i}^{-1} P_{ref,j})^{-1}(P_{est,i}^{-1}P_{est,j})$.}
\begin{center}
\begin{tabular}{l l |c c c}
\multirow{2}{*}{\textbf{Dataset}} & \multirow{2}{*}{\textbf{Model}} &
\multicolumn{3}{c}{\textbf{Evaluate Metrics}}\\
\cline{3-5}
&&APE(m) $\downarrow$ &APE(\%) $\downarrow$ &RPE(m) $\downarrow$\\
\hline
\multirow{4}{*}{KITTI}& ALOAM & 27.05$\pm$13.32 &0.57\%$\pm$0.73\% &1.06$\pm$0.12\\ 
&LeGO-LOAM&170.73$\pm$72.42&4.61\%$\pm$2.62\%&4.50$\pm$0.98\\
&(LiDAR Only)&29.18$\pm$13.17 &0.80\%$\pm$0.84\% & 4.38$\pm$1.13\\ 
&ORBSLAM3&\textbf{5.65$\pm$2.99}&\textbf{0.14\%$\pm$0.11\%}&\textbf{0.91$\pm$0.77} \\  \hline 

\multirow{3}{*}{nuScenes}& ALOAM&27.84$\pm$28.22 &16.41\%$\pm$14.23\%  &0.17$\pm$0.15\\ 
&LeGO-LOAM&\textbf{21.70$\pm$20.65}&\textbf{15.75\%$\pm$15.03\%}&\textbf{0.09$\pm$0.03}\\ 
&ORBSLAM3&\multicolumn{3}{c}{Not Applicable to Our Experiment Settings} \\  
\hline 

\multirow{3}{*}{CARLA}& ALOAM &5.16$\pm$2.85&0.39\%$\pm$0.64\%  &\textbf{0.30$\pm$0.13}\\ 
&LeGO-LOAM&14.44$\pm$9.33&0.99\%$\pm$0.87\%&2.16$\pm$0.48\\ 
&ORBSLAM3 &\textbf{2.70$\pm$1.47}&\textbf{0.18\%$\pm$0.10\%}&0.59$\pm$0.11 \\  \hline 
\end{tabular}
\label{tbl:geometric}
\end{center}
\end{table}
}

\newcommand{\styled}{
\begin{table}[t]
\caption{Performance comparison of learning-based monocular depth estimation models trained on datasets synthetic domain (PreSIL), real-world domain (KITTI) and synthetic dataset processed with SAVeS$^{+}$. For each model, we selected the best checkpoints using SILog during the entire training process. We then used KITTI Depth Prediction Evaluation Dataset to evaluate on three trained models. We used the same evaluation metrics as \tabref{depth}. Best performance is marked in bold, while second best is marked with underline.}
\begin{center}
\scalebox{1.0}{
\begin{tabular}{l l |c c c c}
\multirow{2}{*}{\textbf{Models}} & \multirow{2}{*}{\textbf{Datasets}} & 
\multicolumn{4}{c}{\textbf{Evaluate Metrics}}\\
\cline{3-6}
& &SILog $\downarrow$ &sqErrRel(\%) $\downarrow$ &absErrRel(\%) $\downarrow$ &iRMSE(1/km) $\downarrow$\\ \hline

\multirow{3}{*}{AdaBins}& PreSIL & 1.5790 & 181.5360 &3.2431&0.7419\\ 
&GTA (ours)&\textbf{0.3135}&\textbf{0.2309}&\textbf{0.2651}&\textbf{0.0245}\\
&KITTI&0.5163&0.4835&0.3321&0.0500 \\  \hline 

\multirow{3}{*}{BTS}& PreSIL &0.3402 & 2.4702&0.3076 &0.0244\\ 
&GTA (ours)&\underline{0.2355} &\underline{0.1102} &\underline{0.1934} &\underline{0.0230}\\
&KITTI&\textbf{0.1263} &\textbf{0.0170} &\textbf{0.0841} &\textbf{0.0094} \\  \hline 

\multirow{3}{*}{SwinTransformer}& PreSIL &0.7037 &17.7337 &1.1188 &0.0518\\ 
&GTA (ours)&\underline{0.3870} &\underline{0.3900} &\underline{0.3220} & \underline{0.0542}\\
&KITTI&\textbf{0.1698} &\textbf{0.0320} &\textbf{0.1092} &\textbf{0.0126} \\  \hline
\end{tabular}
}
\label{tbl:styled}
\end{center}
\end{table}
}

\begin{document}


\title{Bridging the Domain Gap between Synthetic and Real-World Data for Autonomous Driving}

\author{Xiangyu Bai} \email{bai.xiang@northeastern.edu}
\author{Yedi Luo}\email{luo.ye@northeastern.edu}
\author{Le Jiang} \email{jiang.l@northeastern.edu}
\author{Aniket Gupta}\email{gupta.anik@northeastern.edu}
\author{Pushyami Kaveti}\email{kaveti.p@northeastern.edu}
\author{Hanumant Singh}\email{ha.singh@northeastern.edu}
\author{Sarah Ostadabbas}\email{ostadabbas@ece.neu.edu}
\affiliation{%
  \institution{Northeastern University}
  \streetaddress{360 Huntington Avenue}
  \city{Boston}
  \state{Massachusetts}
  \country{USA}
  \postcode{02115}
}

\renewcommand{\shortauthors}{Xiang and Yedi, et al.}

\begin{abstract}
Modern autonomous systems require extensive testing to ensure reliability and build trust in ground vehicles. However, testing these systems in the real-world is challenging due to the lack of large and diverse datasets, especially in edge cases. Therefore, simulations are necessary for their development and evaluation. However, existing open-source simulators often exhibit a significant gap between synthetic and real-world domains, leading to deteriorated mobility performance and reduced platform reliability when using simulation data. To address this issue, our Scoping Autonomous Vehicle Simulation (SAVeS) platform benchmarks the performance of simulated environments for autonomous ground vehicle testing between synthetic and real-world domains. Our platform aims to quantify the domain gap and enable researchers to develop and test autonomous systems in a controlled environment. Additionally, we propose using domain adaptation technologies to address the domain gap between synthetic and real-world data with our SAVeS$^+$ extension. Our results demonstrate that SAVeS$^+$ is effective in helping to close the gap between synthetic and real-world domains and yields comparable performance for models trained with processed synthetic datasets to those trained on real-world datasets of same scale. This paper highlights our efforts to quantify and address the domain gap between synthetic and real-world data for autonomy simulation. By enabling researchers to develop and test autonomous systems in a controlled environment, we hope to bring autonomy simulation one step closer to realization\footnote{The simulated in-road datasets, enhanced GTA datasets and SAVeS platform with SAVeS$^+$ plugin is available at \href{https://github.com/ostadabbas/SAVeS-Platform}{https://github.com/ostadabbas/SAVeS-Platform}.}.
\end{abstract}


\received{20 February 2007}
\received[revised]{12 March 2009}
\received[accepted]{5 June 2009}

\maketitle

\section{introduction}
The past decade has seen a significant rise in the popularity of designing and developing autonomous ground vehicles, driven primarily by the goal of improving safety and reliability. To achieve these objectives, modern autonomous systems require continuous evolution through rigorous and thorough testing and evaluations on large-scale datasets \cite{car_not_safe}. However, conducting real-world testing is hindered by high costs and time constraints, and publicly available autonomous driving datasets that cover diverse edge cases are scarce, requiring dedicated hardware, time, and expertise to collect and annotate.

Fortunately, advancements in video game products \cite{gtav} and improved simulators \cite{CARLA} have made it possible to easily acquire synthetic data, resulting in configurable and abundant datasets with diverse scenes and sensor variety \cite{ue4sim}. Synthetic data generation has already demonstrated groundbreaking progress in related fields, including pose estimation \cite{lejiang}, natural language processing \cite{text_syn}, and medical applications \cite{medical}. However, using synthetic data in algorithmic autonomy tasks can lead to significant domain gaps with real-world data, as synthetic data lacks the physical appearance and may exhibit different statistical and geometric properties than those of real-world data. This phenomenon is especially noticeable in visual applications, where the quality of camera imagery is affected by these gaps. In this paper, we aim to quantify and reliably assess the mobility performance of autonomous vehicles under different data domains and subsequently close the gap between synthetic and real-world domains in autonomous applications.

To achieve this goal, we first present a universal test and evaluation platform called Scoping Autonomous Vehicle Simulation (SAVeS), which highlights and quantifies the domain-induced performance gap of autonomy systems developed using in-road synthetic environments versus real-world environments. Our tool compares driving datasets from real versus simulated settings on two standard but crucial autonomous vehicle tasks to quantify the domain gap: monocular depth estimation and trajectory estimation (odometry). The odometry task estimates changes in position over time, while depth estimation measures the distance of each pixel relative to the camera. The performance of the driving dataset on these tasks, represented by their state-of-the-art algorithms, reflects the domain gap between synthetic and real-world territory. We focus on two classes of inference models: \textit{geometric-based} algorithms and \textit{learning-based} models. Geometric-based algorithms reconstruct the 3D world based on computational geometry using features extracted from point clouds and images to localize the vehicle in unknown surroundings, while learning-based models use end-to-end machine learning techniques to learn from training data and directly obtain the desired outputs.

SAVeS enables us to quantify the domain gap between synthetic and real-world datasets by evaluating their performance on state-of-the-art solutions for odometry and depth estimation tasks. To address this domain gap, we further present SAVeS$^{+}$, an extension of the SAVeS platform that incorporates two domain adaptation techniques, namely style transfer and depth mask, into the synthetic data. By doing so, SAVeS$^{+}$ reduces the reliance on real-world data, eliminates the need for expensive professional sensors such as LiDAR and infrared sensors, and avoids tedious calibration and synchronization procedures typical of a real-world autonomous dataset. Moreover, we introduce an enhanced synthetic dataset for monocular depth estimation using our proposed method. We then demonstrate the effectiveness of our pipeline by conducting experiments on several learning-based monocular depth estimation models. Our results show that when evaluated in the real-world domain, models trained on our SAVeS$^{+}$ dataset yield comparable performance to those trained on real-world datasets of same scale.

The contributions of our paper are multi-fold, and can be summarized as follows:
\begin{itemize}
\item We present the Scoping Autonomous Vehicle Simulation (SAVeS) platform, which provides a standardized and universal test and evaluation framework for depth and odometry tasks in the field of autonomous driving. This platform allows for the quantification of the domain gaps between real-world and synthetic datasets, and enables cross-task comparison of autonomous driving solutions.
\item We analyze the performance of state-of-the-art algorithms on various datasets from different domains, and demonstrate the significant disparities between synthetic and real-world data. This highlights the urgent need for domain adaption techniques in order to bridge the gap between synthetic and real-world domains.
\item We introduce SAVeS$^{+}$, an extension of the SAVeS platform that includes a domain adaption pipeline using style transfer and depth mask techniques. This pipeline improves the performance of synthetic autonomy solutions in the real-world domain, and reduces the reliance on expensive professional sensors and tedious calibration procedures.
\item We bring in a novel synthetic autonomy driving dataset that includes essential sensor data such as stereo RGB images, high-frequency inertial data, light detection and ranging sensor (LiDAR) data, and odometry ground truth. This dataset is generated from the CARLA simulator and is suitable for cross-task comparison.
\item We introduce an enhanced synthetic dataset for the monocular depth estimation task, which is adapted to the real-world domain from Grand Theft Auto (GTA V). This dataset is designed to overcome the domain gap between synthetic and real-world data, and yields comparable performance to models trained on real-world datasets.
\end{itemize}
\section{Related Works}
The development of autonomous vehicles requires the successful completion of various tasks, with odometry and depth estimation being two of the most significant and commonly studied ones. These tasks have been tackled using two broad categories of methods: geometric-based algorithms and learning-based models. Despite numerous efforts to improve the accuracy and performance of algorithms for these tasks, relatively little attention has been given to the issue of training data scarcity. In particular, while many studies have focused on improving algorithmic performance using real-world datasets, few have explored the potential of synthetic simulators for addressing the issue of imbalanced data distribution through domain adaptation.
\subsection{Solving Autonomy Tasks}
Simultaneous localization and mapping (SLAM) is a type of geometric-based algorithm that is specifically designed to solve odometry tasks. The output of SLAM is the vehicle trajectory, which indicates the accuracy of the algorithm in reconstructing a 3D environment and tracking the position of the vehicle \cite{odometry}. SLAM approaches can be classified into two categories: LiDAR- and vision-based SLAM. While the industry heavily relies on LiDAR-based SLAM due to its proven accuracy, a more advanced approach is to fuse data from multiple sensors to enhance overall robustness and accuracy. However, this requires external calibrations and may not be easily achievable due to hardware limitations \cite{sensor_fusion}. Depth estimation is a critical task in autonomous vehicle research as it provides essential information for obstacle detection \cite{depth_def}. Various approaches have been proposed for this task, including vision-, sensor-, and learning-based methods. Recently, deep learning techniques have been extensively applied to the monocular depth estimation task \cite{learning_overview}.
\subsection{Autonomous Driving Platforms and Datasets}
Various datasets have been proposed for research on autonomous vehicles, including multi-purpose benchmark suites like KITTI \cite{kitti}, and those developed for specific tasks such as DIODE \cite{diode}. Researchers can also record their own data using highly customizable middleware platforms such as the Robotics Operating System (ROS), which has extensive open-source support. However, most datasets, such as TUM VI \cite{tumvi} and EuRoC \cite{sun_rgbd}, focus primarily on indoor scenarios. Indoor scenes are usually recorded using hand-held devices or small robots, resulting in slow and shaky footage with mostly compact surroundings. Autonomy vehicles, on the other hand, may move rapidly on roads and generate steady footage with visible captures of the sky. These differences can cause disturbances in the results of both geometric-based and learning-based models. Despite the rapidly growing autonomy industry, existing datasets remain inadequate for addressing numerous edge cases and diverse environments. This inadequacy is especially problematic for data-hungry learning-based models. Additionally, the lack of infrastructure and high cost associated with building test vehicles can make data collection and testing difficult. To compensate for this lack of real-world training and testing footage, enterprises and institutes have been utilizing autonomy simulators such as CARLA \cite{CARLA} for a long period. Research agencies have also developed plugins and modifications to driving games such as GTA \cite{gtav} and Microsoft Flight Simulator \cite{airsim}.

\subsection{Domain Adaption and Style Transfer}
Domain adaptation is a form of transfer learning that addresses the issue of differences in feature spaces or distributions between training and testing datasets, with the aim of improving generalization performance \cite{domainadaptionoverview}. In the realm of visual applications, domain adaptation using deep representations has been successfully applied in robotics to address the challenge of low performance when training only on synthetic data \cite{robotadapt}. Style transfer is a specific domain adaptation technique that can be employed to reduce the domain gap. Style transfer involves aligning the distribution of a content image to that of a desired style image \cite{depthstyletf}. In the context of synthetic driving datasets, it is possible to transform the appearance of synthetic environmental objects from their signature animation style to a more realistic style that reflects real-world photographs \cite{vsait}. This enables the use of synthetic images with accurate ground truth, thus avoiding the need to collect and process data from real vehicles and sensors. Training a functional style transfer model requires significantly less image data and computing resources than collecting enough real-world data to train a depth estimation model \cite{lejiang,dpt}. Additionally, existing unpaired style translation algorithms eliminate the need for manual labeling and ground truth annotation.

\section{Methods}

\overviewplus
In this section, we propose SAVeS and SAVeS$^{+}$ to thoroughly quantify and address the domain gap in autonomous datasets (see \figref{overviewplus}). We provide a detailed explanation of the structure and model selection for SAVeS, along with the two modules integrated into SAVeS$^{+}$ to mitigate the domain gap. 

\subsection{SAVeS Platform}
We have developed the Scoping Autonomous Vehicle Simulation (SAVeS) platform for evaluating the mobility performance of autonomous vehicles using both geometric and learning-based algorithms, and real-world and synthetic urban driving datasets (see \figref{overviewplus}). Our pre-configured platform includes a selection of datasets and models, but we have designed it to be modular, allowing for easy swapping of pre-trained models and datasets. The SAVeS platform consists of four modules: the input converter, model containers, output converter, and evaluation components. The input converter resolves issues with dataset format compatibility by re-processing the data into an intermediate format. The model containers and dataset drivers produce estimations, and the output converter translates the estimations to ensure consistency between models and convert the format to match the desired task. Finally, the evaluation components provide comparable results for analysis.




\subsection{Model Selections}
For geometric-based algorithms, we have selected two widely-used LiDAR-based algorithms: Advanced LOAM (ALOAM) \cite{WEBSITE:aloam} and LeGO-LOAM \cite{legoloam}. ALOAM divides the problem into two separate tasks and tackles them independently: a fast but low-fidelity odometry calculation to estimate the velocity of the LiDAR and a low-frequency computation for finer details of the point cloud. On the other hand, LeGO-LOAM is built upon the concept of LOAM and leverages segmentation of LiDAR data to extract ground plane and edge features, and uses a two-step Levenberg-Marquardt optimization to solve them separately.

For vision-based algorithms, we have selected ORBSLAM3 \cite{ORBSLAM3}. This algorithm adds support for the multi-map system with improved place recognition methods. The multi-map system allows the algorithm to tackle visual lost issue by merging existing maps.




In this paper, we evaluate several learning-based models for monocular depth estimation, including GeoNet \cite{geonet},  DPT \cite{dpt}, and AdaBins \cite{adabins}. GeoNet is an unsupervised CNN-based framework that offers easier data collection for training, but it faces gradient locality issues and produces lower output image quality. DPT utilizes vision transformer (ViT) as its backbone, yielding detailed high-quality depth maps with high resolution. However, DPT requires intense computation resources and achieves the best performance only when trained on large-scale datasets. AdaBins proposes adaptive bins, a transformer block that splits the depth interval into adaptively computed bins, but this method underperforms compared to a full transformer trained on very large datasets. We incorporate these learning-based models into our evaluation framework to compare their performance on both synthetic and real-world datasets.

\subsection{SAVeS+ Platform}
SAVeS$^{+}$ is an extension of our SAVeS platform designed to address the domain gap between synthetic and real-world datasets for autonomous driving. It includes a training pipeline for several state-of-the-art depth estimation models and two configurable domain adaptation modules: style transfer and depth masking.

For style transfer, we have incorporated the Vector Symbolic Architecture based image-to-image translation technique (VSAIT) \cite{vsait}. VSAIT uses vector symbolic architectures to learn a shared representation space between the source and target domains. This technique eliminates the need for paired training data, making it possible to adapt one domain to another without manual laboring to produce corresponding ground truth. VSAIT can transfer the training dataset closer to the real-world domain using only several thousand target images, which can be collected from a variety of cameras, including mobile phones.

The depth masking module allows users to adjust the range of depth ground truth in the training dataset. Synthetic datasets contain high-resolution and accurate depth maps, but unlike the real-world, the sky and very far objects also have corresponding depth values in simulators. As a result, the distance ground truth is too large to be comparable to data captured by real-world scientific sensors. Our pipeline allows users to configure the amount of valid depth pixels based on distance to match the data distribution of sensors on a real-world ground vehicle.

With the incorporation of the domain adaptation pipeline in SAVeS$^{+}$, we are able to significantly reduce the domain gap between synthetic and real-world datasets, providing more reliable and accurate performance evaluation for autonomous driving algorithms.

\subsection{Synthetic Datasets}
Our synthetic driving dataset, recorded on the CARLA simulation platform, includes 3 urban scenes featuring different variations of urban environments (city center, urban town, and rural villages), each with an average trajectory length of 1.5km. The dataset is equipped with various sensors, including a pair of front-facing stereo RGB cameras at 10Hz, a synced 64-channel 360$^\circ$ LiDAR, and a 6-axis IMU at 100Hz. It also includes a depth camera, a third-person view, and a GNSS measurement unit with dedicated calibration files. We provide depth ground truth in two forms: a sparse LiDAR projection for depth estimation and accurate depth images for depth completion. The odometry ground truth is obtained directly from the simulator. Our dataset is designed to be compatible with both learning-based and geometric-based algorithms for odometry and depth estimation tasks, thereby enabling cross-task comparison by eliminating domain gaps. 

We also release an enhanced synthetic dataset for monocular depth estimation tasks, derived from GTA V and processed with SAVeS$^{+}$. The dataset contains 18k consecutive RGB in-road urban driving images in KITTI's style, and each 540x360 image is provided with pixel-accurate dense depth map for training. We also included sparse synthetic depth map generated from LiDAR point clouds for evaluation purposes. One can also use SAVeS$^{+}$ to process more synthetic GTA datasets such as MVS-synth\cite{DeepMVS} and GTA-SfM\cite{sfm} to further expand the training data.

\section{Experiments}
We conduct our experiments on the SAVeS and SAVeS$^{+}$ platform, where we evaluate the performance of various models for odometry and depth estimation tasks using synthetic and real-world datasets. Specifically, we compare the performance of our synthetic dataset with real-world data on these tasks. Additionally, we train monocular depth estimation models on datasets processed with our domain adaption pipeline and evaluate their performance on the KITTI real-world dataset \cite{kitti}.

\subsection{Datasets}
 {\bf KITTI} is the most recognized benchmark suite for autonomy platforms and has been adopted universally among the most influential research in this field. It covers a wide variety of autonomy tasks and remains one of the most complete datasets till this day comparing to emerging datasets such as PandaSet\cite{pandaset} and ApolloScape\cite{apolloscape}, which makes it ideal for our cross-task comparisons. 

{\bf nuScenes} \cite{nuscene} is a large-scale autonomous driving dataset that fully shows the dense traffic and challenging driving scenarios in Boston and Singapore. It provides an entire sensor suite of an autonomous vehicle (6 cameras, 1 LiDAR, 5 RADAR, GPS, IMU), which satisfies our data requirements for both tasks. The rich complexity of nuScenes can make up for the lack of environmental complexity and diversity of the KITTI dataset, allowing us to evaluate the data domain differences more comprehensively.
{\bf CARLA} \cite{CARLA} is a state-of-the-art driving simulator that provides the complete sensor suite and ground truth for our needs, thus is the optimal solution for multi-task comparison in synthetic surroundings. CARLA is designed to address the high cost and lack of complexity in existing simulators. It is engineered from the ground up and provides an all-in-one solution, from routing to recording. Being an open-source platform, it enjoys the contribution from the community and supports customized urban scenario set-up; it is prone to bugs and missing features in the main branch, as seen in other open-source software.
{\bf GTA V} was initially released as an open-world game back in 2013. 
Previous contributions in GTA V such as DeepGTAV, and DeepGTA-Presil \cite{gtav} used modified versions of GTA V to collects RGB camera data and depth map data during driving simulation. Accompanied with its unparalleled scene details, traffic accuracy, weather dynamics and the design depth of the physics engine, our test scenes are more diverse in location selection, especially in the simulation of off-road locations. In our depth estimation experiment, we used a subsection of  PreSIL dataset.

\subsection{Implementation Settings}
For depth estimation, we perform a scale-matching algorithm adopted from GeoNet’s evaluation scripts to make prediction scales consistent with each ground truth by multiplying predicted depth with a scalar (the median of ground truth divided by the corresponding median of prediction) for each image:\\
\centerline{\begin{math}y^*=y_{out}\times\dfrac{median(y_{gt})}{median(y_{out})},\end{math}} where \begin{math}
    y_{out}
\end{math} is the raw prediction from the models and \begin{math}
    y_{gt}
\end{math} represents depth ground truth for corresponding images. \begin{math}
    y^*
\end{math} stands for the prediction with matching scales. We then perform linear interpolation to obtain the original size and store predictions into 16-bit PNG files. All three models provide multiple pre-trained weights. 
For GeoNet, both weights are trained on KITTI with two types of splits. We selected Eigen approach for better performance. AdaBins provides weights trained on either NYUv2 \cite{nyuv2} or KITTI and we select NYUv2. For DPT, we choose their medium-scaled hybrid architecture with no fine-tuning. 

For odometry evaluation, we use the origin alignment method within the evo package \cite{evo} to align the estimated and reference trajectories, which allows us to account for drift in our results.

Regarding domain adaptation analysis, we train all models on 18K images with a resolution of 360p from our enhanced GTA dataset. For AdaBins and SwinTransformer, we use their default training parameters, except for setting the batch\_size to 4 instead of 16, and train them on one RTX 3090 for 25 epochs. For BTS, we train the model on 4 Nvidia V100s for 25 epochs with batch\_size=16, using the provided training parameters. For SAVeS$^+$ modules, we randomly select 7000 raw images from the KITTI dataset to train VSAIT style transfer and mask all points with a true depth greater than 80.0, which is the maximum depth in the KITTI dataset. We train all three models on the PreSIL dataset pre-processed with this configuration and evaluate them on 1000 images of the KITTI depth prediction dataset using the scale matching algorithm mentioned before.


\depth
\subsection{Depth Evaluation}
The apparent trend from the results shows that regardless of pre-train datasets, there is a visible domain gap between synthetic imagery and real-world camera footage.

On all models and test datasets, we observe that CARLA footage consistently performs worse than their real-world counterparts by a noticeable margin, with the difference sometimes doubling or even tripling on some metrics. Both relative squared error (sqErrorRel) and absolute error (absErrorRel) metrics align with this trend. However, iRMSE is higher on DPT and AdaBins. This can be attributed to three reasons: 1) most errors occur at the farthest pixels, meaning their values are higher than average; 2) the depth ground truth of real-world datasets is smaller than that in CARLA due to hardware limitations. The mean depth value above the 90th percentile for the CARLA dataset is 73.96, whereas the value in KITTI is 39.42; 3) both DPT and AdaBins tend to have a higher maximum threshold when it comes to synthetic depth prediction.

GeoNet is trained on KITTI, and thus its prediction threshold aligns with the ground truth, resulting in the best performance among the three models on KITTI. GeoNet demonstrates how well a model can perform when domain gaps are minimal (i.e., when using the same dataset). The performance of DPT and AdaBins is heavily impacted since we did not use their fine-tuned versions on KITTI. On CARLA and nuScenes, DPT performs the best, followed by GeoNet and AdaBins. This ranking is expected since DPT (a transformer-based model) has been trained on vast amounts of data, eliminating data saturation, while GeoNet (2018) (although two years older than AdaBins (2020)) was trained on KITTI, an urban dataset featuring on-road scenes similar to our CARLA setting. DPT demonstrates the effectiveness of training on a large volume of data.

Our selection of AdaBins has been trained on NYUv2, an indoor dataset, and thus has the widest domain gap with CARLA. The results reflect the performance degradation under the influence of indoor scenarios.

\geometric

\subsection{Odometry Evaluation}

In summary, our evaluation highlights the advantages and limitations of using synthetic and real-world data for different types of algorithms in the context of autonomous driving. While simulated sensor data can accurately represent environment structures, the level of accuracy may be too high to achieve in real-world sensors. As shown in \tabref{geometric}, sensor-based geometric algorithms prefer synthetic environments due to the high accuracy of the sensor data. However, geometric SLAM models heavily rely on point clouds extracted from sensor data, which can be impacted by sensor calibration errors and unpredictable noise in the manufacturing process. Additionally, some algorithms such as LeGO-LOAM have special requirements for specific types of physical sensors, which can severely impact performance.

In contrast, learning-based methods are not prone to these issues, as they can learn from both synthetic and real-world data without being affected by sensor compatibility issues. However, our evaluation also shows that both vision-based geometric algorithms and machine learning models exhibit domain gaps when applied to synthetic data compared to real-world datasets. For instance, ORBSLAM3 performs slightly worse on synthetic datasets, as shown by the APE (\%) metric, due to the shorter travel distances in the simulated environment. Finally, we note that the nuScenes dataset has minimal overlapping areas with our stereo settings for ORBSLAM3, and the shorter scene length makes algorithms more susceptible to poor sensor data frames and unable to correct themselves with loop closure.

\styled
\quality
\subsection{Domain Adaption Evaluation}
Experiment results shown in \tabref{styled} indicates that our proposed pipeline in SAVeS$^{+}$ can greatly improve the generalization performance of models trained on synthetic datasets in real-world testing, meanwhile, yields complete and rich depth estimations. We trained three learning-based models using PreSIL dataset processed with our pipeline, namely AdaBins, BTS\cite{bts} and SwinTransformer\cite{mim}. 

Quantitatively speaking, in AdaBins, the model trained with our processed GTA data from PreSIL dataset reach the same accuracy level as the model trained with KITTI dataset under same image number on KITTI depth validation set, indicating that by removing the domain gap, high-quality synthetic dataset can unleash the full potential and demonstrate the added benefit of various scenes and complex modeling of the simulation surroundings. One reason why the original PreSIL dataset performs poorly is that GTA's full render distance can be over 10,000 meters, contributing to the accumulation of error across the pixels. In both BTS and SwinTransformer, we can observe a noticeable improvement in terms of evaluation performance when training with our processed PreSIL dataset over the original PreSIL dataset. Synthetic dataset processed using SAVeS$^{+}$ yields training depth distribution and range much closer to the targeted real-world dataset and such range can be learned by the inference models and  reflected in the evaluation results. This phenomenon is indicatied by metrics that incorporated absolute value ranges as well as distributions, namely sqErrorRel and absErrorRel, where a significant improvement can be observed.

There is another advantage of training with synthetic data that regrettably cannot be captured by quantitative metrics. Firstly, the KITTI development kit masks all depth pixels greater than 80.0, making dense synthetic depth information irrelevant. Secondly, KITTI uses sparse depth from LiDAR as ground truth, leading to only pixels within the appropriate range being taken into account, which affects both training and validation. Models trained and evaluated on KITTI exhibit the best performance in terms of evaluation metrics due to the absence of domain gaps. However, they fail to produce a dense, clear depth map. Despite the wider depth range deteriorating evaluation accuracy, all three models benefited from it and the dense ground truth. Both the original PreSIL dataset and our processed dataset demonstrate the ability to clearly differentiate sky and very far objects (see \figref{quality}). Using human vision, we can distinguish most objects from the predicted depth images, which is not possible for models trained with the KITTI dataset.

\section{Conclusion and Future Work}
This paper presents a comprehensive effort to address the domain gap between synthetic and real-world datasets for autonomy tasks. The SAVeS platform is introduced to assess the performance of odometry and depth estimation algorithms against each other on real-world and simulated autonomy datasets. The experimental results show that the current open-sourced simulators are not optimized solely to provide real-world details or reflect physical dynamics accurately, and this leads to degraded testing performance of simulated data. To address this issue, we introduce SAVeS$^{+}$ and demonstrate its effectiveness in closing the domain gap and improving the testing performance of simulated data using domain adaptation techniques. Our work has practical implications for the development of autonomous vehicles and related applications. By providing a platform to evaluate and compare the performance of algorithms on synthetic and real-world datasets, we can better understand the strengths and limitations of different approaches and make more informed decisions. Furthermore, our efforts to close the domain gap using domain adaption techniques can help improve the accuracy and reliability of autonomous systems, especially in scenarios where data collection is difficult or expensive. In the future, we plan to extend our platform by including more models and algorithms, as well as additional business-grade simulators. We also aim to explore the use of SAVeS$^{+}$ to help improve geometric-based algorithms on synthetic datasets. Our work demonstrates the importance of bridging the gap between synthetic and real-world data for autonomous systems, and we hope that our efforts will inspire further research in this area.

\section{Acknowledgement}
Research was sponsored by the DEVCOM Analysis Center and was accomplished under Cooperative Agreement Number W911NF-22-2-0001. The views and conclusions contained in this document are those of the authors and should not be interpreted as representing the official policies, either expressed or implied, of the Army Research Office or the U.S. Government. The U.S. Government is authorized to reproduce and distribute reprints for Government purposes notwithstanding any copyright notation herein.

\newpage
{\small
\bibliographystyle{IEEEbib}
\bibliography{refs}
}
\end{document}